\documentclass[11pt]{article}
\usepackage[left=1in,right=1in,top=1in,bottom=1in]{geometry}
\usepackage{times}
\usepackage{hyperref}
\usepackage{graphicx}
\usepackage{amsmath}
\usepackage{listings}
\usepackage{xcolor}
\usepackage{cleveref}
\hypersetup{
  colorlinks   = true, 
  urlcolor     = blue, 
  linkcolor    = blue, 
  citecolor   = red 
}

\begin{document}
\begin{center}
{\huge OpenAI Gym }
\end{center}
\centerline{
Greg Brockman,
Vicki Cheung,
Ludwig Pettersson,}
\centerline{
Jonas Schneider,
John Schulman,
Jie Tang,
Wojciech Zaremba
}
\centerline{\textit{OpenAI}}
\begin{abstract}
OpenAI Gym\footnote{\href{gym.openai.com}{gym.openai.com}} is a toolkit for reinforcement learning research.
It includes a growing collection of benchmark problems that expose a common interface, and a website where people can share their results and compare the performance of algorithms. This whitepaper discusses the components of OpenAI Gym and the design decisions that went into the software.
\end{abstract}

\section{Introduction}
Reinforcement learning (RL) is the branch of machine learning that is concerned with making sequences of decisions. 
RL has a rich mathematical theory and has found a variety of practical applications \cite{bertsekas1995dynamic}.
Recent advances that combine deep learning with reinforcement learning have led to a great deal of excitement in the field, as it has become evident that general algorithms such as policy gradients and Q-learning can achieve good performance on difficult problems, without problem-specific engineering \cite{Mnih15,Schulman15TRPO,mnih2016asynchronous}.

To build on recent progress in reinforcement learning, the research community needs good benchmarks on which to compare algorithms.
A variety of benchmarks have been released, such as the Arcade Learning Environment (ALE) \cite{Bellemare13ALE}, which exposed a collection of Atari 2600 games as reinforcement learning problems, and recently the RLLab benchmark for continuous control \cite{duan2016benchmarking}, to which we refer the reader for a survey on other RL benchmarks, including \cite{RLPy,RLGlue,PyBrain,RLLibCapital,dimitrakakis2014reinforcement}.
OpenAI Gym aims to combine the best elements of these previous benchmark collections, in a software package that is maximally convenient and accessible.
It includes a diverse collection of tasks (called \textit{environments}) with a common interface, and this collection will grow over time.
The environments are versioned in a way that will ensure that results remain meaningful and reproducible as the software is updated.

Alongside the software library, OpenAI Gym has a website (\href{gym.openai.com}{gym.openai.com}) where one can find scoreboards for all of the environments, showcasing results submitted by users.
Users are encouraged to provide links to source code and detailed instructions on how to reproduce their results.

\section{Background}
Reinforcement learning assumes that there is an agent that is situated in an environment. Each step, the agent takes an \textit{action}, and it receives an \textit{observation} and \textit{reward} from the environment. 
An RL algorithm seeks to maximize some measure of the agent's total reward, as the agent interacts with the environment.
In the RL literature, the environment is formalized as a partially observable Markov decision process (POMDP) \cite{Sutton98RLBook}.

OpenAI Gym focuses on the episodic setting of reinforcement learning, where the agent's experience is broken down into a series of \textit{episodes}.
In each episode, the agent's initial state is randomly sampled from a distribution, and the interaction proceeds until the environment reaches a terminal state.
The goal in episodic reinforcement learning is to maximize the expectation of total reward per episode, and to achieve a high level of performance in as few episodes as possible.

\lstset{
language=Python,
basicstyle=\ttfamily,
otherkeywords={self},             
keywordstyle=\ttfamily\color{blue!90!black},
keywords=[2]{True,False},
keywords=[3]{ttk},
keywordstyle={[2]\ttfamily\color{yellow!80!orange}},
keywordstyle={[3]\ttfamily\color{red!80!orange}},
emph={MyClass,__init__},          
emphstyle=\ttfamily\color{red!80!black},    
stringstyle=\color{green!80!black},
commentstyle=\color{gray},
showstringspaces=false            
}

The following code snippet shows a single episode with 100 timesteps.
It assumes that there is an object called {\tt agent}, which takes in the observation at each timestep, and an object called {\tt env}, which is the environment.
OpenAI Gym does not include an agent class or specify what interface the agent should use; we just include an agent here for demonstration purposes.

\small
\begin{lstlisting}
    ob0 = env.reset() # sample environment state, return first observation
    a0 = agent.act(ob0) # agent chooses first action
    ob1, rew0, done0, info0 = env.step(a0) # environment returns observation, 
    # reward, and boolean flag indicating if the episode is complete.
    a1 = agent.act(ob1)
    ob2, rew1, done1, info1 = env.step(a1)
    ...
    a99 = agent.act(o99)
    ob100, rew99, done99, info2 = env.step(a99)
    # done99 == True  =>  terminal
\end{lstlisting}

\section{Design Decisions} \label{dd}

The design of OpenAI Gym is based on the authors' experience developing and comparing reinforcement learning algorithms, and our experience using previous benchmark collections.
Below, we will summarize some of our design decisions.

{\flushleft \textbf{Environments, not agents}.
Two core concepts are the agent and the environment. We have chosen to only provide an abstraction for the environment, not for the agent.
This choice was to maximize convenience for users and allow them to implement different styles of agent interface.
First, one could imagine an ``online learning'' style, where the agent takes {\tt (observation, reward, done)} as an input at each timestep and performs learning updates incrementally.
In an alternative ``batch update'' style, a agent is called with observation as input, and the reward information is collected separately by the RL algorithm, and later it is used to compute an update.
By only specifying the agent interface, we allow users to write their agents with either of these styles.
}

{\flushleft \textbf{Emphasize sample complexity, not just final performance}.
The performance of an RL algorithm on an environment can be measured along two axes: first, the final performance; second, the amount of time it takes to learn---the sample complexity.
To be more specific, final performance refers to the average reward per episode, after learning is complete.
Learning time can be measured in multiple ways, one simple scheme is to count the number of episodes before a threshold level of average performance is exceeded.
This threshold is chosen per-environment in an ad-hoc way, for example, as 90\% of the maximum performance achievable by a very heavily trained agent.

Both final performance and sample complexity are very interesting, however, arbitrary amounts of computation can be used to boost final performance, making it a comparison of computational resources rather than algorithm quality.

}

{\flushleft \textbf{Encourage peer review, not competition}.
The OpenAI Gym website allows users to compare the performance of their algorithms.
One of its inspiration is \href{http://kaggle.com}{Kaggle}, which hosts a set of machine learning contests with leaderboards.
However, the aim of the OpenAI Gym scoreboards is not to create a competition, but rather to stimulate the sharing of code and ideas, and to be a meaningful benchmark for assessing different methods.

RL presents new challenges for benchmarking.
In the supervised learning setting, performance is measured by prediction accuracy on a test set, where the correct outputs are hidden from contestants.
In RL, it's less straightforward to measure generalization performance, except by running the users' code on a collection of unseen environments, which would be computationally expensive. Without a hidden test set, one must check that an algorithm did not ``overfit'' on the problems it was tested on (for example, through parameter tuning).

We would like to encourage a peer review process for interpreting results submitted by users.
Thus, OpenAI Gym asks users to create a \textit{Writeup} describing their algorithm, parameters used, and linking to code.
Writeups should allow other users to reproduce the results.
With the source code available, it is possible to make a nuanced judgement about whether the algorithm ``overfit'' to the task at hand.

}

{\flushleft \textbf{Strict versioning for environments}.
If an environment changes, results before and after the change would be incomparable.
To avoid this problem, we guarantee than any changes to an environment will be accompanied by an increase in version number.
For example, the initial version of the CartPole task is named \texttt{Cartpole-v0}, and if its functionality changes, the name will be updated to \texttt{Cartpole-v1}.

{\flushleft \textbf{Monitoring by default}.
By default, environments are instrumented with a \texttt{Monitor}, which keeps track of every time \texttt{step} (one step of simulation) and \texttt{reset} (sampling a new initial state) are called.
The Monitor's behavior is configurable, and it can record a video periodically. It also is sufficient to produce learning curves.
The videos and learning curve data can be easily posted to the OpenAI Gym website.

\begin{figure}
  \centering
      \includegraphics[height=0.18\linewidth]{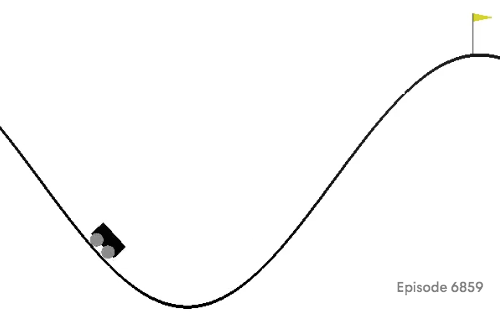}    
      \includegraphics[height=0.18\linewidth]{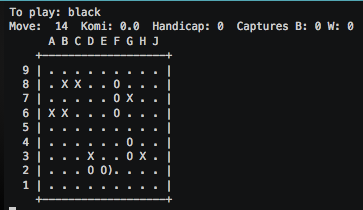}        
      \includegraphics[height=0.18\linewidth]{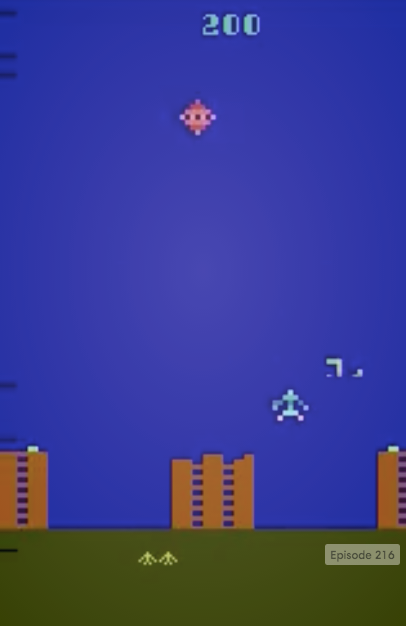} 
      \includegraphics[height=0.18\linewidth]{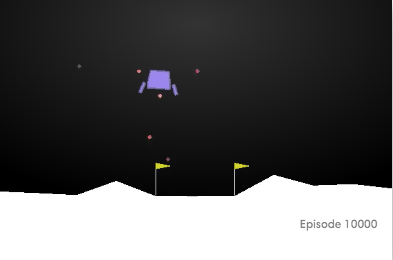}   \\
      \includegraphics[height=0.18\linewidth]{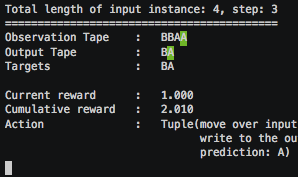}        
      \includegraphics[height=0.18\linewidth]{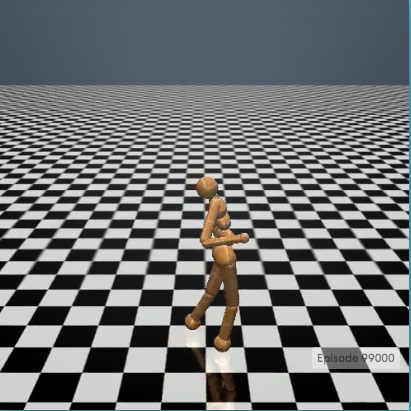}         
      \includegraphics[height=0.18\linewidth]{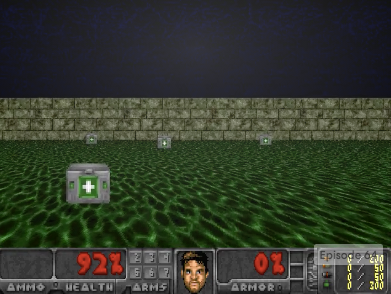}      
      \includegraphics[height=0.18\linewidth]{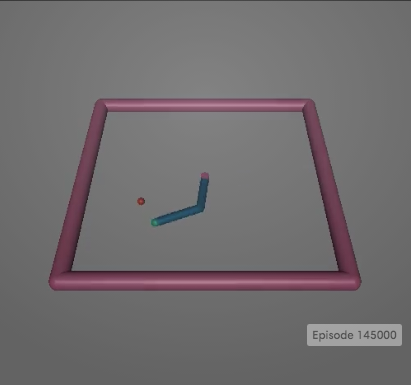}            
   \caption{Images of some environments that are currently part of OpenAI Gym.}
   \label{fig:envs}
\end{figure}

\section{Environments}
OpenAI Gym contains a collection of Environments (POMDPs), which will grow over time.
See \Cref{fig:envs} for examples.
At the time of Gym's initial beta release, the following environments were included:
\begin{itemize}
\item \textit{Classic control and toy text}: small-scale tasks from the RL literature.
\item \textit{Algorithmic}: perform computations such as adding multi-digit numbers and reversing sequences. Most of these tasks require memory, and their difficulty can be chosen by varying the sequence length.
\item \textit{Atari}: classic Atari games, with screen images or RAM as input, using the Arcade Learning Environment \cite{Bellemare13ALE}.
\item \textit{Board games}: currently, we have included the game of Go on 9x9 and 19x19 boards, where the Pachi engine \cite{baudivs2011pachi} serves as an opponent.
\item \textit{2D and 3D robots}: control a robot in simulation. These tasks use the MuJoCo physics engine, which was designed for fast and accurate robot simulation \cite{todorov2012mujoco}.
A few of the tasks are adapted from RLLab \cite{duan2016benchmarking}.
\end{itemize}

Since the initial release, more environments have been created, including ones based on the open source physics engine Box2D or the Doom game engine via VizDoom \cite{kempka2016vizdoom}.

\section{Future Directions}
In the future, we hope to extend OpenAI Gym in several ways.
\begin{itemize}
  \item \textit{Multi-agent setting}. It will be interesting to eventually include tasks in which agents must collaborate or compete with other agents.
  \item \textit{Curriculum and transfer learning}. Right now, the tasks are meant to be solved from scratch. Later, it will be more interesting to consider sequences of tasks, so that the algorithm is trained on one task after the other. Here, we will create sequences of increasingly difficult tasks, which are meant to be solved in order.
  \item \textit{Real-world operation}. Eventually, we would like to integrate the Gym API with robotic hardware, validating reinforcement learning algorithms in the real world.
\end{itemize}

\bibliographystyle{unsrt}
\bibliography{gym-whitepaper}
\end{document}